%% file: main.tex
\documentclass[runningheads]{llncs}
\newcommand{\orcidmanuel}	{\href{https://orcid.org/0000-0002-8510-1972}{\protect\includegraphics[scale=0.05]{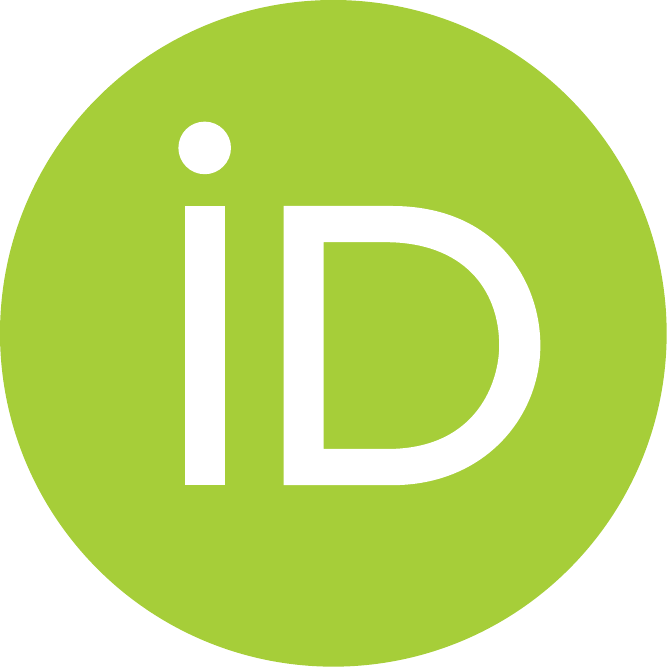}}}
\newcommand{\orcidoscar}	{\href{https://orcid.org/0000-0002-8296-6620}{\protect\includegraphics[scale=0.05]{images/orcid}}}
\newcommand{\orcidmarlon}	{\href{https://orcid.org/0000-0002-9247-7476}{\protect\includegraphics[scale=0.05]{images/orcid}}}
\usepackage{makeidx}  
\usepackage{graphicx}
\usepackage{booktabs} 
\usepackage{comment}
\usepackage{float}
\usepackage{soul}
\usepackage[table,xcdraw]{xcolor}
\usepackage{csquotes}
\usepackage[ruled]{algorithm2e} 

\usepackage{subcaption}
\usepackage[T1]{fontenc}
\usepackage{hyperref}
\usepackage{cleveref}
\usepackage{cite}
\usepackage{multirow}
\usepackage[table,xcdraw]{xcolor}
\usepackage[normalem]{ulem}
\useunder{\uline}{\ul}{}
\usepackage{booktabs}

\begin{document}

%
\title{Discovering Generative Models from Event Logs: Data-driven Simulation vs Deep Learning}
\titlerunning{Data-driven simulation vs deep learning}
%
%
\author{Manuel Camargo\inst{1,2}~\orcidmanuel \and Marlon Dumas\inst{1}~\orcidmarlon \and Oscar Gonz{\'a}lez-Rojas\inst{2}~\orcidoscar}

\authorrunning{M. Camargo et al.}
%
\institute{University of Tartu, Tartu, Estonia, \email{\{manuel.camargo, marlon.dumas\}@ut.ee}
\and
Universidad de los Andes, Bogot{\'a}, Colombia, \email{o-gonza1@uniandes.edu.co}
}

\maketitle              

\begin{abstract}
A generative model is a statistical model that is able to generate new data instances from previously observed ones. In the context of business processes, a generative model creates new execution traces from a set of historical traces, also known as an event log. Two families of generative process simulation models have been developed in previous work: data-driven simulation models and deep learning models. Until now, these two approaches have evolved independently and their relative performance has not been studied. This paper fills this gap by empirically comparing a data-driven simulation technique with multiple deep learning techniques, which construct models are capable of generating execution traces with timestamped events. The study sheds light into the relative strengths of both approaches and raises the prospect of developing hybrid approaches that combine these strengths.

 \end{abstract} 

%
\input{tex/section1}

%
\input{tex/section2}

\input{tex/section3}

%
\input{tex/section4}
%
\input{tex/section5}
%
\input{tex/section6}
%
\bibliographystyle{splncs04}
\bibliography{bib/references}
\end{document}

%% file: tex/section1.tex
\section{Introduction}
\label{sec:intro}

An event log is a collection of execution traces of a business process. Each trace in a log consists of a sequence of events and each event consists of a process instance (case) identifier, an activity label, an activity start and end timestamp, and possibly also the resource who performed the activity and other attributes.

A generative model of a business process is a statistical model constructed from an event log, which is able to generate traces that resemble those observed in the log as well as other traces of the process. Generative process models have several applications, including anomaly detection~\cite{Nolle2018}, predictive monitoring~\cite{Tax2017}, and what-if analysis~\cite{Camargo2020}. Two families of generative models have been studied in the literature: Data-Driven Simulation (DDS) and Deep Learning (DL) models.


DDS models are discrete-event simulation models constructed from an event log. Several authors have proposed techniques for discovering DDS models, ranging from semi-automated techniques~\cite{Martin2016} to  automated ones~\cite{Rozinat2009,Camargo2020}. A DDS model is generally constructed by first discovering a process model from an event log and then fitting a number of parameters (e.g. mean inter-arrival rate, branching probabilities, etc.) in a way that maximizes the similarity between the traces that the DDS model generates and those in (a subset of) the event log. 

On the other hand, DL generative models are machine learning models consisting of  interconnected layers of artificial neurons adjusted based on input-output pairs in order to maximize accuracy. Generative DL models have been widely studied in the context of predictive process monitoring~\cite{Tax2017,Evermann2017,lin2019mm, taymouri2020predictive}, where they are used to generate the remaining path (suffix) of an incomplete trace by repeatedly predicting the next event. It has been shown that these models can also be used to generate entire traces \cite{Camargo2019}.


To date, the relative accuracy of these two families of generative process models has not been studied, barring a study that compares DL models vs automatically discovered process models that generate events without timestamps~\cite{Tax2018}. This paper fills this gap by empirically comparing these approaches using five event-logs with varying structural and temporal characteristics. Based on the evaluation results, the paper discusses the relative strengths and potential synergies of these approaches.

The paper is organized as follows. Sections 2 and 3 review DDS and DL generative modeling approaches, respectively, and introduce the approaches included in the evaluation. Section 4 presents the evaluation setup, while Section 5 discusses the results. Finally, Section 6 concludes and outlines future work.

%% file: tex/section2.tex
\section{Data-driven simulation}

Business Process Simulation (BPS) is a quantitative process analysis technique in which a discrete-event model of a process is stochastically executed a number of times, and the resulting simulated execution traces are used to compute aggregate performance measures such as the average waiting times of activities or the average cycle time of the process~\cite{FBPM}. 

Typically, a BPS model consists of a \emph{process model} enhanced with time and resource-related parameters such as the the \emph {inter-arrival time} of cases and its associated Probability Distribution Function (PDF), the PDFs of each activity's processing times, a \emph{branching probability} for each conditional branch in the process model, and the \emph{resource pool} responsible for performing each activity type in the process model~\cite{FBPM}. Such BPS models are stochastically executed by creating new cases according to the inter-arrival time PDF, and simulating the execution of each case following the control-flow semantics of the process model and the following activity execution rules: (i) If an activity in a case is enabled, and there is an available resource in the pool associated to this activity, the activity is started and allocated to one of the the available resources in the pool; (ii) When the completion time of an activity is reached, the resource allocated to the activity is made available again. Hence, the waiting time of an activity is entirely determined by the availability of a resource. Resources are assumed to be eager: as soon as a resource is assigned to an activity, the activity is started. 



A key ingredient for BPS is the availability of a BPS model that accurately reflects the actual dynamics of the process. Traditionally, BPS models are created manually by domain experts by gathering data from interviews, contextual inquiries, and on-site observation. In this approach, the accuracy of the BPS model is limited by the accuracy of the process model used as a starting point. 

Several techniques for discovering BPS models from event logs have been proposed~\cite{Martin2016, Rozinat2009}. These approaches automate the extraction of the process model from an event log, and then enhance this model with all the simulation parameters derived from the event log (e.g.\ arrival rate). In this paper, we use the term DDS model to refer to a BPS model discovered from an event log.

Existing approaches for discovering a DDS from an event log can be classified in two categories. The first category consists of approaches that provide conceptual guidance to discover BPS models. For example, \cite{Martin2016} discusses how PM techniques can be used to extract, validate, and tune BPS model parameters, without seeking to provide fully automated support. Similarly, \cite{Wynn2008} outlines a series of steps to construct a DDS model using process mining techniques.

The second category of approaches seek to automate the extraction of simulation parameters. For example, \cite{Rozinat2009} proposes a pipeline for constructing a DDS using process mining techniques. However, in this approach, the tuning of the simulation model (i.e..\ fitting the parameters to the data) is left to the user. 

\begin{figure}[ht]
  \begin{center}
    \includegraphics[width=\textwidth]{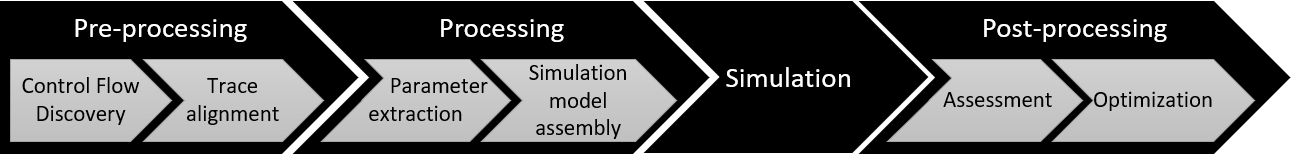}
    \caption{Pipeline of Simod to generate process models.}
    \label{fig:architecture}
   \end{center}
   \vspace{-8mm}
\end{figure}

In this paper, we use a DDS discovery method, namely Simod tool~\cite{Camargo2020}, which automates both the construction and the tuning of the DDS model.  
Simod combines several PM techniques to automate the generation, and validation of BPS models from an event log. Fig.~\ref{fig:architecture} illustrates the steps of the Simod method: Pre-processing, Processing, Simulation, and Post-processing.

In the \textbf{Pre-processing} stage, Simod extracts a BPMN model from data and guarantees its quality and coherence with the event log. The first step is the \textit{Control Flow Discovery}, using the SplitMiner algorithm~\cite{Augusto2017}, which is known for being one of the fastest, simple, and accurate discovery algorithms. Next, Simod applies \textit{Trace alignment} to assess the conformance between the discovered process model and each trace in the input log. The tool provides options for handling non-conformant traces, via removal, replacement, or repair, to ensure obtain full conformance, which is needed in the following stages.

In the \textbf{Processing} stage, Simod extracts the simulation parameters and assembles them in a single BPS model. The extracted parameters correspond to the \textit{Resource pools} involved in the process, the probability density function (PDF) of \textit{Inter-arrival times} and \textit{Activities durations}, and the \textit{branching probabilities}. The resource pool is discovered using the algorithm proposed by Song and Van der Aalst~\cite{Song2008}; likewise, the resources are assigned to the different activities according to the frequency of execution. The inter-arrival times and activities durations PDFs are discovered by fitting a collection of possible distribution functions to the data series, selecting the one that yields the smallest standard error. For calculating the branching probabilities, the tool offers two options: assign equal values to each conditional branch or compute the traversal frequencies of the conditional branches by replaying the event log on the process model. Finally, once compiled all the simulation parameters, these are assembled with the BPMN model into a single data structure that can be interpreted by a discrete event simulator (e.g., Bimp) in the \textbf{Simulation} step.


In the \textbf{Post-processing} stage, Simod assesses the similarity between the event log generated by a simulation and the ground truth log, using a measure of similarity between event logs (ELS), which we introduce in Section 4. Simod then uses a Bayesian hyperparameter optimizer to explore the search space of all possible Simod parameter settings, in search for a configuration of parameters that leads to the highest ELS between the simulated log and a ground-truth log.



%% file: tex/section3.tex
\section{Generative deep learning models of business processes}
\label{subsec:dl}

In recent years, the use of generative deep learning models has been widely studied in the field of predictive process monitoring. Multiple proposals~\cite{Evermann2017, Tax2017, Camargo2019} have demonstrated that such models achieve high accuracy for tasks such as predicting the next event of a running case (and its timestamp or other attribbutes such as the resource) as well as predicting the remaining path of an incomplete case. 

In broad terms, a Deep Learning (DL) model is a network composed of multiple interconnected layers of neurons (perceptrons), which perform non-linear transformations of data~\cite{Hao2016}. The main goal of these transformations is training the network to learn the behaviors/patterns observed in the data. Theoretically, the more layers of neurons there are in the system, the more it becomes possible to detect higher-level patterns in the data thanks to the composition of complex functions~\cite{Lecun2015}. In the literature multiple neural network architectures (e.g., feed-forward, convolutional, autoencoders, etc.) have been used in domains such as natural language processing or image processing. In the field of predictive process monitoring the most common type of neural network is the Recurrent Neural Networks (RNN) due to the sequential nature of the input event logs.

In particular, Evermann et al.~\cite{Evermann2017} proposed an approach to generate the most likely remaining sequence of events (suffix) starting from a prefix of an ongoing case. However, this architecture cannot handle numerical features, and hence it cannot generate sequences of timestamped events. The approaches of Lin et al. and Taymouri et al.~\cite{lin2019mm, taymouri2020predictive} also shares this inability to predict timestamps and durations. An alternative approach by Tax et al.~\cite{Tax2017} can predict timestamps; however, it lacks flexibility in the management of high dimensional inputs due to one-hot-encode categorical features. As a result, its accuracy deteriorates as the number of categorical features increases. In \cite{Tax2018}, the same authors compare the performance of several techniques for predicting the next element in a sequence using real-life datasets. This latter study addresses the problem of predicting the next event's type, but it does not consider the problem of simultaneously predicting the next event and its timestamp. Finally, Nolle et al~\cite{Nolle2018} propose a neural network architecture called BINet for real-time anomaly detection in business process executions. 

In this paper, we use the DeepGenerator method proposed in~\cite{Camargo2019} to train generative DL models for the task of generating complete event logs. 

\begin{figure}[ht]
  \centering
    \includegraphics[width=.8\textwidth]{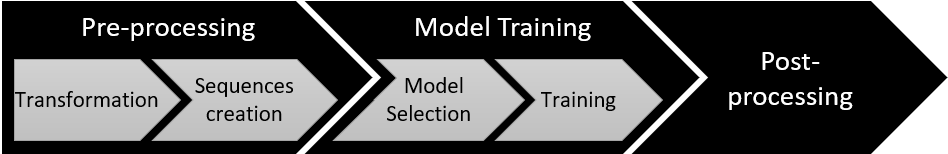}
    \caption{Phases and steps for building DL models}
    \label{fig:steps}
   \vspace*{-3mm}
\end{figure}

The DeepGenerator method extends previous approaches for training DL models with LSTM architectures, by including dimensionality control techniques such as the use of n-grams and embedded dimensions, as well as the exploration of random sampling following probability distribution for the category selection of the next predicted event. Since this method was designed to generate events with a single timestamp, we extended it in this paper to support the prediction of two timestamps. Similarly, we generalized the method and use it to train other kind of RNN architecture known as GRU to broaden the scope of the evaluation performed in this paper. Fig.~\ref{fig:steps} synthesizes the phases and stages for building predictive models with our method.

In the \textbf{Pre-processing phase}, we carry out a \textit{Transformation} of the event log by generating n-sized sequences of events composed of activities, roles, and times. Here, we use encoding and scaling techniques depending on the data type of the event attribute (i.e., categorical, or continuous). In the case of the \textit{categorical attributes}, i.e., activities and roles, they were encoded using the embedded dimensions technique. This method helps us to keep the dimensionality low, which enhances the performance of the neural network. The embedded dimensions are n-dimensional spaces in which the models can map the activities and roles according to their proximity. In the case of \textit{continuous attributes}, that is, start and end timestamps, they are first relativized and later scaled over a range of $[0, 1]$. The relativization is carried out by first calculating two features: the event duration and the time-between-events. The duration of an event (a.k.a.\ the processing time) is the time difference between its end and its start timestamp. The time-between-events is the difference between the event's start and the end of the immediately preceding event in the same trace (a.k.a.\ the waiting time). All relative times are scaled using normalization and log-normalization depending on the variability of the times in the event log. Once the features are encoded, the next step is the \textit{Sequences creation} step. In this step, we extract n-grams from each trace to create the example sequences to train the predictive models. One n-gram is generated for each step of the process execution and this is done for each attribute independently. Hence, we use four independent inputs: activity prefixes, role prefixes, relativized event durations, and relativized time-between-events. 

\begin{figure}[ht]
    \centering
    \begin{subfigure}[t]{0.30\textwidth}
        \includegraphics[width=\textwidth]{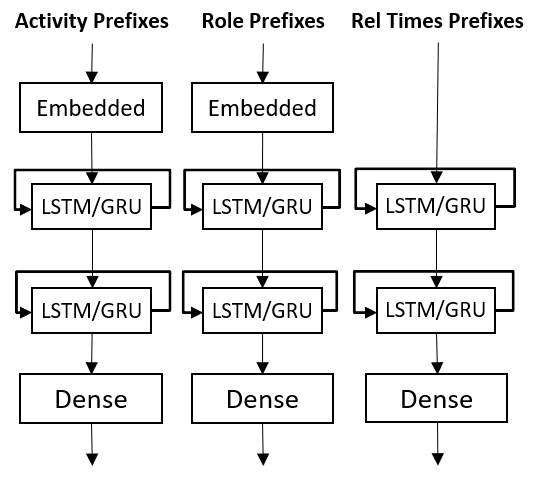}
        \caption{Specialized}
        \label{fig:spec}
    \end{subfigure}
    \begin{subfigure}[t]{0.30\textwidth}
        \includegraphics[width=\textwidth]{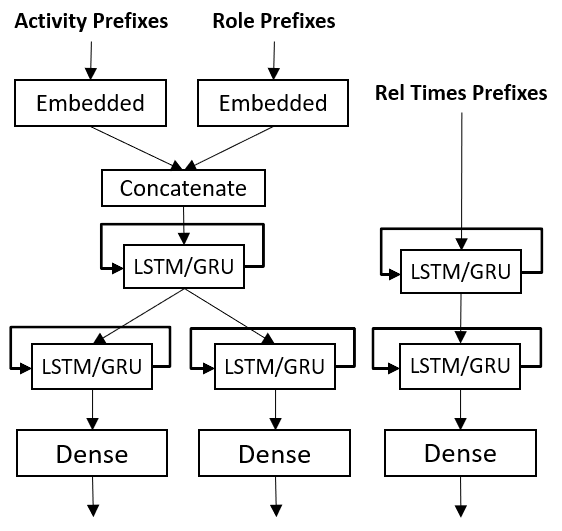}
        \caption{Shared categorical}
        \label{fig:shared}
    \end{subfigure}
    \begin{subfigure}[t]{0.30\textwidth}
        \includegraphics[width=\textwidth]{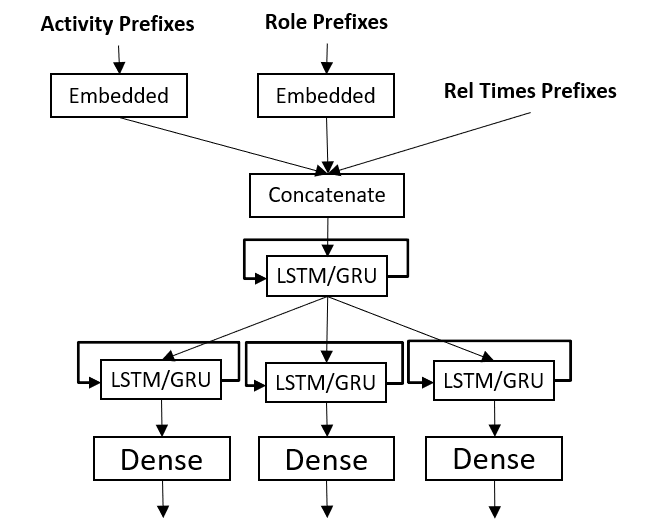}
        \caption{Full shared}
        \label{fig:full}
    \end{subfigure}
    \caption{(a) this architecture does not share any information, (b) this architecture concatenates the inputs related with activities and roles, and shares the first layer, (c) this architecture completely shares the first layer.}
    \label{fig:architechtures}    
    \vspace{-3mm}
\end{figure}

In the \textbf{Model Training Phase}, one of three possible stacked base architectures is selected for training. The network structures vary according to whether or not they share intermediate LSTM or GRU layers, considering that sometimes sharing information can help to differentiate execution patterns. Fig.~\ref{fig:architechtures} presents the general structure of the defined architectures.

Finally, the \textbf{Post-processing Phase} includes the mechanisms for generating complete traces from a zero-prefix size. The way of doing this is using continuous feedback of the model with each newly generated event, until the generation of a finalization event. The category of the next event is selected randomly following the predicted probability distribution. This mechanism turns out to be the most suitable for the task of generating complete event logs by avoiding getting stuck in the higher probabilities, as was tested in \cite{Camargo2019}.

%% file: tex/section4.tex
\section{Evaluation}
\label{sec:evaluation}

This section presents an empirical comparison of DDS and DL generative process models. The evaluation aims at addressing the following questions: what is the relative accuracy of these approaches when it comes to generating traces of events without timestamps? and what is their relative accuracy when it comes to generating traces of events with timestamps?

\subsection{Datasets}



We evaluated the selected approaches using five event logs that contain both start and end timestamps:
\begin{itemize}
    \item The event log of a manufacturing production (MP) process contains the steps exported from an Enterprise Resource Planning (ERP) system~\cite{production_log}. 
    \item The event log of a purchase-to-pay (P2P) process is a synthetic log generated from a model not available to the authors.\footnote{The log is provided as part of the  Fluxicon Disco tool -- \url{https://fluxicon.com/}} 
    \item The event log from an Academic Credentials Recognition (ACR) process of a Colombian University was gathered from its BPM system (Bizagi). 
    \item The W subset of the BPIC2012\footnote{\url{https://doi.org/10.4121/uuid:3926db30-f712-4394-aebc-75976070e91f}} event log, which is a log of a loan application process from a Dutch financial institution. The W subset of this log is composed of the events corresponding to activities performed by human resources (i.e.\ only activities that have a duration). 
    \item The W subset of the  BPIC2017\footnote{\url{https://doi.org/10.4121/uuid:5f3067df-f10b-45da-b98b-86ae4c7a310b}} event log, which is an updated version of the the BPIC2012 log. We carried out the extraction of the W-subset by following the recommendations reported by the winning teams participating in the BPIC 2017 challenge \footnote{\url{https://www.win.tue.nl/bpi/doku.php?id=2017:challenge}}.
    \end{itemize}

Table~\ref{tab:eventlogs} summarizes the characteristics of these logs. The BPIC2017W and BPIC2012W logs have the largest number of traces and events, while the MP, P2P and ACR have less traces but a higher average number of events per trace.
\begin{table}[ht]
\centering
\scriptsize
\resizebox{\textwidth}{!}{%
\begin{tabular}{@{}llllllll@{}}
\toprule
\textbf{Event log} & \textbf{\begin{tabular}[c]{@{}c@{}}Num. \\ traces\end{tabular}} & \textbf{\begin{tabular}[c]{@{}c@{}}Num. \\ events\end{tabular}} & \textbf{\begin{tabular}[c]{@{}c@{}}Num. \\ activities\end{tabular}} & \textbf{\begin{tabular}[c]{@{}c@{}}Avg. \\ activities \\ per trace\end{tabular}} & \textbf{\begin{tabular}[c]{@{}c@{}}Max. \\ activities \\ per trace\end{tabular}} & \textbf{\begin{tabular}[c]{@{}c@{}}Mean \\ duration\end{tabular}} & \textbf{\begin{tabular}[c]{@{}c@{}}Max. \\ duration\end{tabular}} \\ \midrule
\multicolumn{1}{l|}{MP} & \multicolumn{1}{l|}{225} & \multicolumn{1}{l|}{4953} & \multicolumn{1}{l|}{26} & \multicolumn{1}{l|}{22} & \multicolumn{1}{l|}{177} & \multicolumn{1}{l|}{20.6 days} & 87 days 10 hours \\ \midrule
\multicolumn{1}{l|}{P2P} & \multicolumn{1}{l|}{608} & \multicolumn{1}{l|}{9119} & \multicolumn{1}{l|}{21} & \multicolumn{1}{l|}{14.9} & \multicolumn{1}{l|}{44} & \multicolumn{1}{l|}{21.5 days} & 108 days 7 hours \\ \midrule
\multicolumn{1}{l|}{ACR} & \multicolumn{1}{l|}{954} & \multicolumn{1}{l|}{6870} & \multicolumn{1}{l|}{18} & \multicolumn{1}{l|}{7.2} & \multicolumn{1}{l|}{23} & \multicolumn{1}{l|}{14.9 days} & 135 days 19 hours \\ \midrule
\multicolumn{1}{l|}{BPIC2012W} & \multicolumn{1}{l|}{8616} & \multicolumn{1}{l|}{59302} & \multicolumn{1}{l|}{6} & \multicolumn{1}{l|}{6.88} & \multicolumn{1}{l|}{74} & \multicolumn{1}{l|}{8.9 days} & 85 days 20 hours \\ \midrule
\multicolumn{1}{l|}{BPIC2017W} & \multicolumn{1}{l|}{30276} & \multicolumn{1}{l|}{240854} & \multicolumn{1}{l|}{8} & \multicolumn{1}{l|}{7.9} & \multicolumn{1}{l|}{65} & \multicolumn{1}{l|}{12.7 days} & 286 days 1 hour \\ \bottomrule
\end{tabular}%
}
\caption{Event logs description}
\label{tab:eventlogs}
\end{table}


\subsection{Evaluation measures}

To measure the accuracy of a generative process model, we use it to generate an event log (multiple times) and we measure the average similarity between the generated logs and a ground-truth event log. To this end, we define four measures of similarity between pairs of logs: Control-Flow Log Similarity (CFLS), Mean Absolute Error (MAE) of cycle times, Event Log Similarity (ELS), and Earth-Mover's Distance (EMD) of the histograms of activity processing times.

CFLS is defined based on a measure of distance between pairs of traces: one trace coming from  the original event log and the other from the generated log. We first convert each trace into a sequence of activity (i.e.\ we drop the timestamps and other attributes). In this way, a trace becomes a sequence of symbols (i.e.\ a string). We then measure the difference between two traces using the Damerau-Levenshtein distance, which is the minimum number of edit operations necessary to transform one string (a trace in our context) into another. The supported edit operations are insertion, deletion, substitution, and transposition. Transpositions are allowed without penalty when two activities are concurrent, meaning that they appear in any order, i.e.\ given two activities, we observe both AB and BA in the log. Next, we normalize the resulting Damerau-Levenshtein distance by dividing the number edit operations by the length of the longest sequence. We then define the \emph{control-flow trace similarity} as the one minus the normalized Damerau-Levenshtein distance. Given this trace similarity notion, we pair each trace in the generated log with a trace in the original log, in such a way that the sum of the trace similarities between the paired traces is maximal. This pairing is done using the Hungarian algorithm for computing optimal alignments \cite{Kuhn1955}. Finally, we define the CFLS between the real and the generated log as the average similarity of the optimally paired traces.


The \emph{cycle time MAE} measures the temporal similarity between two logs. The absolute error of a pair of traces T1 and T2 is the absolute value of the difference between the cycle time of T1 and that of T2. The cycle time MAE is the mean of the absolute errors over a collection of paired traces. Like for the CFLS measure, we use the Hungarian algorithm to pair each trace in the generated log with a corresponding trace in the original log.

The cycle time MAE is a rough measure of the temporal similarity between the traces in the original and the generated log. It does not take into account the timing of  the events in a trace -- only the cycle time of the full trace. To complement the cycle time MAE, we use the Earth Mover's Distance (EMD) between the normalized histograms of the mean durations of the activities in the ground-truth log vs the same histogram computed from the generated log. The EMD between two histograms H1 and H2 is the minimum number of units that need to be added to, removed to, or transferred across columns in H1 in order to transform it into H2. The EMD is zero if the observed mean activity durations in the two logs are identical, and it tends to one the more they differ.

The above measures focus either on the control-flow or on the temporal perspective. To complement them, we use the a measure that combines both perspective, namely the ELS as defined in~\cite{Camargo2020}. This measure is defined in the same way as CLFS above, except that it uses a distance measure between traces that takes into account both the activity labels and the timestamps of activity labels.
This distance measure between traces is called Business Process Trace Distance (BPTD). The BPTD measures the distance between traces composed of events that occur in time intervals. This metric is an adaptation of the CFLS metric that, in the case of label matching, assigns a penalty based on the differences in times. BPTD also supports parallelism, which commonly occurs in business processes. We have called ELS to the generalization of the BPTD to measure the distance between two event logs using the Hungarian algorithm.
\vspace*{-3mm}

\subsection{Experiment setup} 

\begin{figure}[ht]
  \centering
    \includegraphics[width=0.7\textwidth]{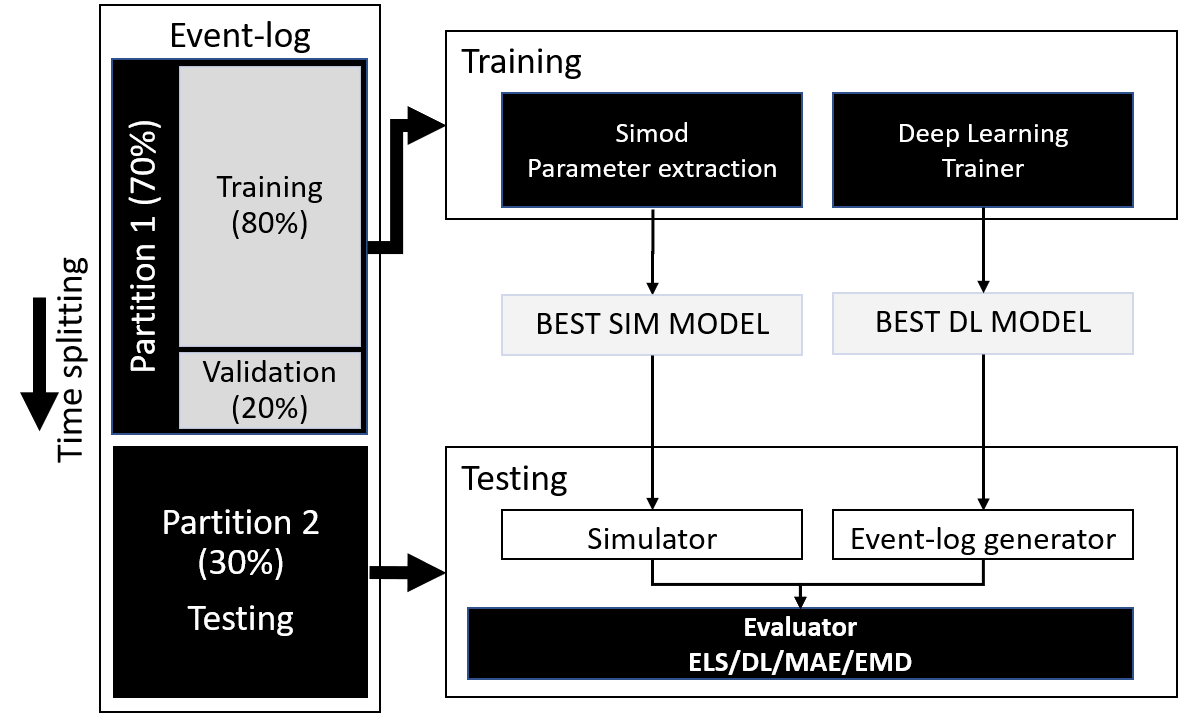}
    \caption{Experimental pipeline}
    \label{fig:exp_pipeline}
    \vspace*{-3mm}
\end{figure}

The aim of the evaluation is to compare the accuracy of DDL models vs DL models discovered from event logs. 
Fig.~\ref{fig:exp_pipeline} presents the pipeline we followed.

We used the holdup method with a temporal split criterion to divide the event logs into two folds: 70\% for training and 30\% for testing. Next, we use the training fold to train the DDS and the DL models. 
We use Simod's hyperparameter optimizer to tune the DDS model. The optimizer is set to explore 50 parameter configurations with five simulation runs per configuration, using the first 80\% of the training fold to construct candidate DDS models and the remaining 20\% for validation. We retained the DDS model that gave the best results on the validation sub-fold in terms of ELS averaged across the five runs. 
Next, for each model family (LSTM and GRU) we apply random search for hyperparameter optimization. Like in the DDS approach, we explore 50 random configurations with five runs each, using 80\% of the training fold for model construction and 20\% for validation. The above led us to one DDS, one LSTM and one GRU model per log.  We then generated ten logs per retained model. To ensure comparability, each generated log was of the same size (number of traces) as the testing fold of the original log. For each generated log, we then compare it to the testing fold using the ELS, CFLS, EMD and MAE measures defined above. To smooth out stochastic variations, we report the mean of each of these measures across the 10 logs generated from each model.


%% file: tex/section5.tex
\section{Results and Discussion} 
\label{sec:results}

Table~\ref{tab:results} presents the evaluation results. The \emph{Event-log} column identifies the evaluated log, the column \emph{Model Family} refers to the type of model (DDS, LSTM, GRU). The ELS, CFLS, MAE and EMD columns present the accuracy measures. Note that ELS and CFLS are similarity measures (higher is better) whereas MAE and EMD are error/distance measures (lower is better).

\begin{table}[ht]
\scriptsize
\centering
\resizebox{0.80\textwidth}{!}{%
\begin{tabular}{@{}lcccccc@{}}
\toprule
\textbf{Type of log} & \textbf{Event log} & \textbf{Model family} & \textbf{ELS} & \textbf{CFLS} & \textbf{MAE} & \textbf{EMD} \\ \midrule
\multicolumn{1}{l|}{\multirow{3}{*}{Synthetic}} & \multicolumn{1}{c|}{\multirow{3}{*}{P2P}} & \multicolumn{1}{c|}{GRU} & \multicolumn{1}{c|}{0.63} & \multicolumn{1}{c|}{{\ul \textit{\textbf{0.58}}}} & \multicolumn{1}{c|}{2551993.36} & {\ul \textit{\textbf{9.02}}} \\ \cmidrule(l){3-7} 
\multicolumn{1}{l|}{} & \multicolumn{1}{c|}{} & \multicolumn{1}{c|}{LSTM} & \multicolumn{1}{c|}{{\ul \textit{\textbf{0.70}}}} & \multicolumn{1}{c|}{0.56} & \multicolumn{1}{c|}{{\ul \textit{\textbf{2310094.83}}}} & 9.90 \\ \cmidrule(l){3-7} 
\multicolumn{1}{l|}{} & \multicolumn{1}{c|}{} & \multicolumn{1}{c|}{DDS} & \multicolumn{1}{c|}{0.62} & \multicolumn{1}{c|}{0.57} & \multicolumn{1}{c|}{2689451.20} & 10.11 \\ \midrule
\multicolumn{1}{l|}{\multirow{3}{*}{Real}} & \multicolumn{1}{c|}{\multirow{3}{*}{MP}} & \multicolumn{1}{c|}{GRU} & \multicolumn{1}{c|}{0.34} & \multicolumn{1}{c|}{0.37} & \multicolumn{1}{c|}{{\ul \textbf{243375.60}}} & 2.81 \\ \cmidrule(l){3-7} 
\multicolumn{1}{l|}{} & \multicolumn{1}{c|}{} & \multicolumn{1}{c|}{LSTM} & \multicolumn{1}{c|}{0.35} & \multicolumn{1}{c|}{0.36} & \multicolumn{1}{c|}{447143.67} & 2.98 \\ \cmidrule(l){3-7} 
\multicolumn{1}{l|}{} & \multicolumn{1}{c|}{} & \multicolumn{1}{c|}{DDS} & \multicolumn{1}{c|}{{\ul \textit{\textbf{0.44}}}} & \multicolumn{1}{c|}{{\ul \textit{\textbf{0.42}}}} & \multicolumn{1}{c|}{488158.75} & {\ul \textit{\textbf{2.10}}} \\ \midrule
\multicolumn{1}{l|}{\multirow{3}{*}{Real}} & \multicolumn{1}{c|}{\multirow{3}{*}{ACR}} & \multicolumn{1}{c|}{GRU} & \multicolumn{1}{c|}{0.58} & \multicolumn{1}{c|}{0.58} & \multicolumn{1}{c|}{{\ul \textit{\textbf{354516.66}}}} & {\ul \textit{\textbf{2.45}}} \\ \cmidrule(l){3-7} 
\multicolumn{1}{l|}{} & \multicolumn{1}{c|}{} & \multicolumn{1}{c|}{LSTM} & \multicolumn{1}{c|}{0.56} & \multicolumn{1}{c|}{0.56} & \multicolumn{1}{c|}{369663.50} & 2.58 \\ \cmidrule(l){3-7} 
\multicolumn{1}{l|}{} & \multicolumn{1}{c|}{} & \multicolumn{1}{c|}{DDS} & \multicolumn{1}{c|}{{\ul \textit{\textbf{0.84}}}} & \multicolumn{1}{c|}{{\ul \textit{\textbf{0.87}}}} & \multicolumn{1}{c|}{430365.05} & 5.35 \\ \midrule
\multicolumn{1}{l|}{\multirow{3}{*}{Real}} & \multicolumn{1}{c|}{\multirow{3}{*}{BPI2012W}} & \multicolumn{1}{c|}{GRU} & \multicolumn{1}{c|}{0.49} & \multicolumn{1}{c|}{0.49} & \multicolumn{1}{c|}{723567.90} & 183.47 \\ \cmidrule(l){3-7} 
\multicolumn{1}{l|}{} & \multicolumn{1}{c|}{} & \multicolumn{1}{c|}{LSTM} & \multicolumn{1}{c|}{0.55} & \multicolumn{1}{c|}{0.55} & \multicolumn{1}{c|}{{\ul \textit{\textbf{333609.01}}}} & 189.03 \\ \cmidrule(l){3-7} 
\multicolumn{1}{l|}{} & \multicolumn{1}{c|}{} & \multicolumn{1}{c|}{DDS} & \multicolumn{1}{c|}{{\ul \textit{\textbf{0.57}}}} & \multicolumn{1}{c|}{{\ul \textit{\textbf{0.58}}}} & \multicolumn{1}{c|}{745528.89} & {\ul \textit{\textbf{162.14}}} \\ \midrule
\multicolumn{1}{l|}{\multirow{3}{*}{Real}} & \multicolumn{1}{c|}{\multirow{3}{*}{BPI2017W}} & \multicolumn{1}{c|}{GRU} & \multicolumn{1}{c|}{0.83} & \multicolumn{1}{c|}{{\ul \textit{\textbf{0.83}}}} & \multicolumn{1}{c|}{984222.14} & {\ul \textit{\textbf{279.30}}} \\ \cmidrule(l){3-7} 
\multicolumn{1}{l|}{} & \multicolumn{1}{c|}{} & \multicolumn{1}{c|}{LSTM} & \multicolumn{1}{c|}{{\ul \textit{\textbf{0.85}}}} & \multicolumn{1}{c|}{0.75} & \multicolumn{1}{c|}{{\ul \textit{\textbf{678099.59}}}} & 289.73 \\ \cmidrule(l){3-7} 
\multicolumn{1}{l|}{} & \multicolumn{1}{c|}{} & \multicolumn{1}{c|}{DDS} & \multicolumn{1}{c|}{0.50} & \multicolumn{1}{c|}{0.47} & \multicolumn{1}{c|}{1008748.01} & 547.26 \\ \bottomrule
\end{tabular}%
}
\caption{Evaluation results}
\label{tab:results}
\end{table}


Both DDS and DL approaches gave similar results on the artificial log (P2P). This may be due to the fact that this log has a predictable behavior, which is equally well captured by both families of approaches. In three of the four real-life logs (MP, ACR, and BPI2012W), the DDS model has higher accuracy in terms of control-flow similarity, particularly in the ACR log. However, in the BPI2017W log, the DL models yielded considerably higher control-flow similarity. This can be explained by the fact that this log is much larger than the others, and DL models generally excel when fed large amounts of samples. Turning our attention to temporal similarity, we note that DL models led to the best MAE results across all event logs. An example of this can be observed in the BPI2012W log in which the best generative model obtained half the MAE than the best simulation model. In the case of EMD, there is no clear winner. All approaches are able to accurately reproduce the distribution of processing times of activities. 

The results indicate that DDS models perform well when it comes to capturing the occurrence and order of activities (control-flow), event with smaller training datasets. However, the Deep Learning models are more accurate when it comes to capturing the temporal perspective and, as expected, they perform particularly well for the largest dataset. 

A possible explanation is that event logs of business processes (at least the ones included in this evaluation) follow to certain normative pathways that can be captured sufficiently well by automatically discovered process models. On the other hand, the waiting times in these event logs are not adequately captured by DDS models. 
DL models on the other hand are able to find patterns in the observed event timestamps. Since the EMD values are similar for DDS and DL models, we conclude that the differences in temporal accuracy between these two types of models stems from the fact that DL models are better able to predict the waiting times of activities (rather than the processing times). 
The inability for DDS models to accurately capture the waiting times can be attributed to the fact that these models rely on the assumption that the waiting times can be fully explained by the availability of resources (i.e.\ resource contention is the sole cause of waiting times) and that they operate under the assumption of \emph{eager resources} as  discussed in Section 2 (i.e.\ resources start an activity as soon as it is allocated to them). DL models on the other hand simply try to find the best possible fit to the observed waiting times. 


The results of this paper are restricted to the five event-logs used for the evaluation. To obtain results with a considerable statistical significance is needed a larger volume of event logs with the required characteristics. Similarly, this work does not include all possible DL architectures applied in the context of business processes; the results are restricted to LSTM and GRU models.
\vspace*{-3mm}

%% file: tex/section6.tex
\section{Conclusion} \label{sec:conclusiones}

In this paper, we compared the accuracy of two approaches to discover generative models from event logs: Data-Driven Simulation (DDS) and Deep Learning (DL). The results suggest that DDS models are suitable for capturing the sequence of activities (or possibly other categorical attributes) of a process. On the other hand, DL models outperform DDS models when predicting the timing of activities, specifically the waiting times between activities.

This observation raises the prospect of combining these approaches into hybrid techniques that take advantage of their relative strengths. In such hybrid approaches, the DDS model would capture the control-flow perspective, while the DL model would capture the temporal dynamics, particularly waiting times. The DSS model would also provide an interpretable model that users can change in order to define ``what-if'' scenarios, e.g. a what-if scenario where an activity is removed or a new activity is added. The challenges here are: (i) how to integrate the DDS model with the DL model; and (ii) how to incorporate the information in a what-if scenario into a DL model. A possible direction to tackle the latter challenge is to adapt existing techniques to incorporate domain knowledge (e.g.\ the fact that an activity has been deleted) into the output of a DL model~\cite{DBLP:conf/bpm/Francescomarino17}.


\medskip\noindent\textbf{Reproducibiltiy package} The scripts and datasets required to reproduce the reported evaluation can be found at: \url{https://github.com/AdaptiveBProcess/DDSvsDL}

\medskip\noindent\textbf{Acknowledgments} Research supported by the European Research Council (PIX project).